%% file: main.tex
\documentclass{acm_proc_article-sp}

\usepackage{url}
\usepackage{multirow}
\usepackage{todonotes}
\usepackage{algorithm,algorithmic}

\begin{document}

%\title{Cross-Document Named Entity Co-reference \\ and Clustering: a Virtuous Cycle}
\title{Joint Event Detection and Entity Resolution: \\ a Virtuous Cycle}

%\subtitle{[Extended Abstract]
%\titlenote{A full version of this paper is available as
%\textit{Author's Guide to Preparing ACM SIG Proceedings Using
%\LaTeX$2_\epsilon$\ and BibTeX} at

\numberofauthors{3} %  in this sample file, there are a *total*
\author{
% 1st. author
\alignauthor Matthias Gall\'e \\
       \affaddr{Xerox Research Centre Europe}\\
       \affaddr{6 Chemin de Maupertuis}\\
       \affaddr{38240 Meylan, France }\\
       \email{matthias.galle@xrce.xerox.com}
% 2nd. author
\alignauthor Jean-Michel Renders \\
       \affaddr{Xerox Research Centre Europe}\\
       \affaddr{6 Chemin de Maupertuis}\\
       \affaddr{38240 Meylan, France }\\
       \email{jean-michel.renders@xrce.xerox.com}
% 3rd. author
\alignauthor Guillaume Jacquet \\
       \affaddr{Xerox Research Centre Europe}\\
       \affaddr{6 Chemin de Maupertuis}\\
       \affaddr{38240 Meylan, France }\\
       \email{guillaume.jacquet@xrce.xerox.com}
}
\maketitle
\begin{abstract}
Clustering web documents has numerous applications, such as aggregating news articles into meaningful events, detecting trends and hot topics on the Web, preserving diversity in search results, etc. 
At the same time, the importance of named entities and, in particular, the ability to recognize them and to solve the associated co-reference resolution problem are widely recognised as key enabling factors when mining, aggregating and comparing content on the Web. 

Instead of considering these two problems separately, we propose in this paper a method that tackles jointly the problem of clusteirng news articles into events and cross-document co-reference resolution of named entities. 
The co-occurrence of named entities in the same clusters is used as an additional signal to decide whether two referents should be merged into one entity.
These refined entities can in turn be used as enhanced features to re-cluster the documents and then be refined again, entering into a virtuous cycle that improves simultaneously the performances of both tasks.
We implemented a prototype system and report results using the TDT5 collection of news articles, demonstranting the potential of our approach.
\end{abstract}

%A category including the fourth, optional field follows...
%\category{D.2.8}{Software Engineering}{Metrics}[complexity measures, performance measures]
\category{H.2.8}{Database Applications}{Data Mining}
\category{H.3.3}{Information Search and Retrieval}{Clustering}
\category{I.2.7}{Natural Language Processing}{Text analysis}
\category{I.7}{Document and Text Processing}{Document Preparation}

\terms{Event-based information systems, natural-language processing, clustering}

\keywords{Co-reference resolution, entity linking, topic detection and tracking, news aggregator} % NOT required for Proceedings

\section{Introduction}

\input{introduction.tex}

\section{Related Work}
\label{sec:priorart}

\input{related.tex}

\section{Joint Event Recognition and Entity Resolution}
\label{sec:corefer}
\input{method.tex}

%\section{Feedback for Clustering}
%\label{sec:feedback}
%\input{feedback.tex}

\section{Experiments and Discussion}
\label{sec:experiments}

\input{experiments.tex}

\section{Conclusions}
\label{sec:conclusion}
\input{conclusions.tex}

\bibliographystyle{abbrv}
\bibliography{biblio}  % sigproc.bib is the name of the Bibliography in this case
% ACM needs 'a single self-contained file'!
%\balancecolumns

\end{document}

%% file: introduction.tex
Behind real-life web-oriented data mining applications are in most cases a suite of different modules, patched together in a sensible way. 
However, this ecosystem of algorithms is seldom acknowledged in research papers, which mostly focus on solving one single task, ignoring the general picture.
On the one hand, this combination of algorithms can of course affect negatively the performance of the whole system, as errors tend to accumulate and serve as catalyzers of other errors. 
But on the other hand, this could also be used to boost the effectiveness of a single task, by resolving it together with another. 
Let as an example a news article aggregator, the application that originated this research.
Modern news aggregators combine multiple Natural Language Processing (NLP) components such as news article stemming/lemmatizing and parsing, Named Entity Recognition (NER), co-reference resolution, document clustering into news events\footnote{In this paper, by ``event'' we mean ``set of documents (or piece of documents) that refer to the same precise topic''}, sentiment analysis, \textit{etc.}. 
These components are executed independently, or in a pipeline where one component waits for the output from the other(s) in (for example) the order we listed above.
Using a NER component and a co-reference resolution component for the clustering is straightforward as the named entities can be used as additional useful features when computing similarity measure between documents.
In this paper, we analyze how the document clustering itself can improve the resolution of the cross-document named entity co-reference, which in turn is used to improve the clustering performance.
This process can then be repeated, entering into a virtuous cycle that improves the resolution of both tasks.

Cross-document named entity co-reference consists in identifying in a set of documents all the textual expressions that refer to the same real entity referent (henceforth "entity" in this paper). It can be split into two sub-tasks, each with its own set of challenges. 
Intra-document co-reference resolution is more concerned with pronouns resolution, implicit mentions and linking correctly different noun phrases.
Inter-document co-reference resolution on its side considers a set of documents and has to harmonize different or incomplete textual expressions referring to the same entity. It has to resolve spelling variants, use of parts of the complete name, acronyms, etc. 

Cross-document named entity co-reference is particularly crucial in Information Extraction systems and namely in news event tracking, where new named entities/entity referents appear all the time and where there is a special interest in relating events through the actors (people and organizations) that took part in them and through the locations they occur in.
Such news event tracking systems however have an advantage that has not been yet exploited in the literature. 
It creates clusters of news articles talking about specific events, providing social\footnote{in this paper, by "social co-occurrences" we mean co-occurrence of entities in the same event, i.e. in the same cluster of news articles}, geographical and temporal co-occurrences of the entities mentioned in the articles. 
This makes it possible to relate different textual expressions used by different sources for the same entity, which would otherwise slip through a standard named entity co-reference system.

\begin{figure}
 \includegraphics[width=250pt]{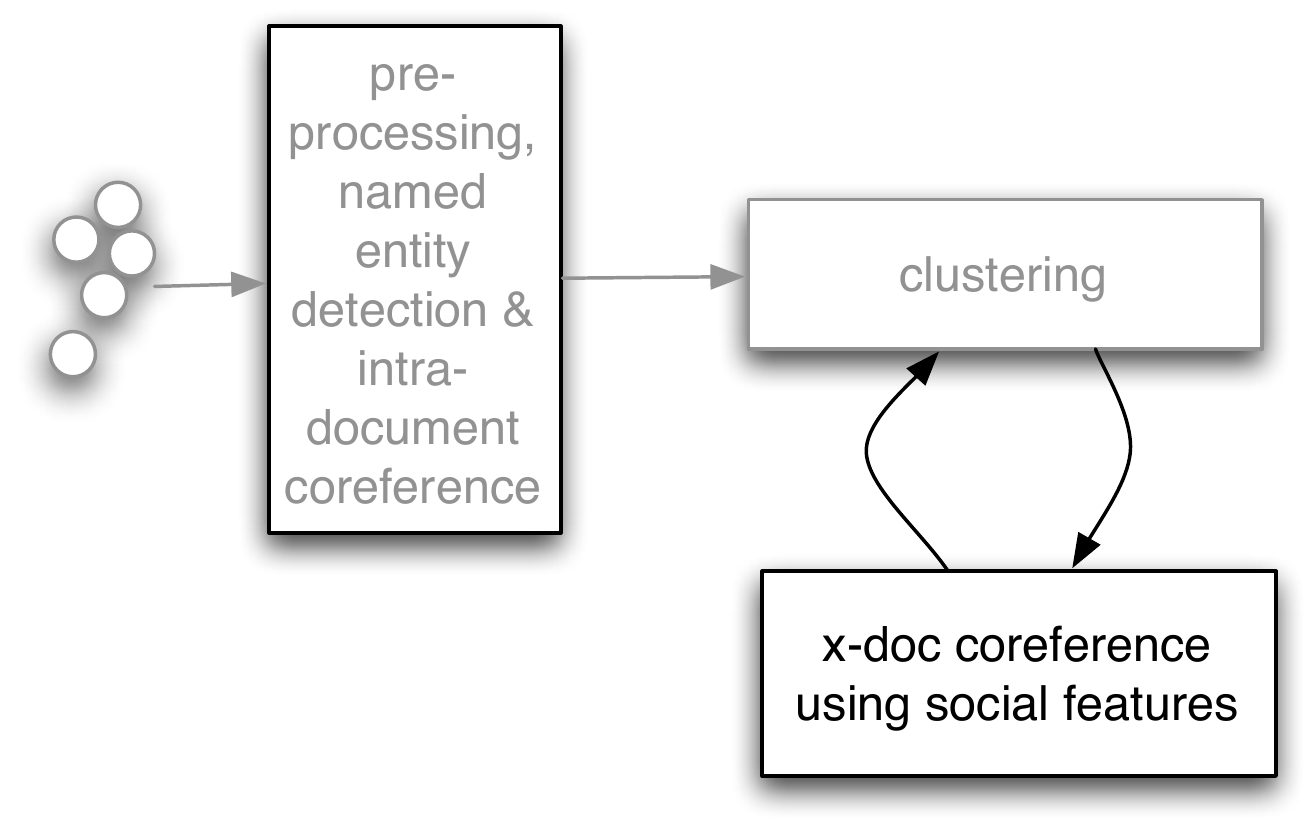}
\caption{An overview of the system we propose. 
The documents arrive (in a stream, mini-batches or full-batch) and are preprocessed; named entities are extracted and intra-document co-reference resolution is applied. 
For the initial cross-document co-reference resolution, a trivial approach is used (two entities co-refer if there is an exact match between their canonical strings). 
Finally the documents are clustered and the information of co-occurrences in events is then used for an enhanced cross-document co-referencing. 
This is then fed back to the clustering algorithm and several iterations can be performed (fixed number or until convergence). 
Our novelty lies in the bold parts (the bootstrap cycle and the co-reference box).}
\label{fig:outline}
\end{figure}

We propose the use of co-occurrences in events as additional features (to be used in combination with other more traditional features, such as string similarities on the canonical surface form) in order to determine if two named entities co-refer to the same underlying entity. 
We use this building block in a system where this enhanced co-reference information can then be fed back to the clustering module in order to improve its quality.
A general outline of such a system is depicted in Fig.~\ref{fig:outline}.
It receives as input a stream of documents.
These are first preprocessed (html to text cleaning, lemmatization, \textit{etc.}) and a NER system is run on each document individually. 
This NER system includes intra-document co-reference resolution, therefore the output is, for each document, a list of entities -- represented by a \textit{canonical string} -- together with the number of times this entity appears (whatever the textual expression used) inside the document. The module is generally able to determine a canonical string, which is non-ambiguous, i.e. the most specific ones with respect to all mentions of an entity in a document. 
Such a intra-document co-reference component misses of course any possible variants due to spelling errors, transcription differences, \textit{etc.}, besides those due to parser's errors (such as taking a form $\langle$ADJ$\rangle$ $\langle$NOUN$\rangle$ as a named entity, e.g. "Faraway Andalusia").
Experimentally, we observed that state-of-the-art intra-document co-reference resolution algorithms very often tend to prioritize precision over recall, and over-generate entities.
The phenomenon of co-reference is therefore much more common than the one of entity ambiguity, and our cross-document co-reference module addresses this by deciding which entities should be merged. 
In other words, we are considering the named entity co-reference problem as an alias detection task: we suppose that a unique real entity has several aliases (in particular because different sources of information use different aliases) and that, by mining the socio-temporal behavior of these aliases, we are able to merge two (or more) aliases as unique entity.

The first contribution of this paper is the use of social, geographical and temporal information in order to merge the named entities that co-refer.
The second contribution is the creation of a virtuous cycle between the clustering and the cross-document co-reference module in order to iteratively improve the quality of both tasks.

%% file: related.tex
This work stands in the intersection of two research fields in data mining, namely Topic Detection and Tracking (TDT) and cross-document co-reference resolution.
While we did not find any previous work on tackling these two problems jointly, there is a rich literature considering these problems separately.

TDT~\cite{Allan1998} is concerned with monitoring news providers in order to extract \textit{events}, which group news articles reporting the same concrete and precise topic.
The most basic algorithm here is an incremental clustering algorithm that compares newly arrived articles to existing events and decides to assign it to one of the existing ones (Topic Tracking) or to create a new one (New Topic Detection). 
$k-$based clustering algorithms, where the number of clusters are pre-defined are unsuitable for this tasks due to the dynamic, constantly evolving nature of the newsphere. 
Several advanced clustering techniques have been proposed for this~\cite{Vadrevu2011,Galle12} which instead take a similarity threshold as parameter, which controls how tight or sparse the cluster should be (corresponding to different levels of granularity of the events).
In this work, we use the fully incremental (or single-pass) algorithm~\cite{Rijsbergen} that, while not the most effective algorithm for this task~\cite{Galle12}, runs fast and allows us to measure the impact of different parameters.
This algorithm processes the document one by one in chronological order.
The first document creates a singleton cluster.
A subsequent document is compared to the centroids of the existing clusters and the highest such similarity is recorded.
If this similarity is above a pre-fixed threshold $\tau$ then the document gets assigned to this cluster and its centroids is updated.
If not, it creates a singleton cluster by its own.

Classical works using statistical methods for named entity co-reference resolution traditionally compute a similarity between two mentions of an entity in the text, and then join some of these mentions into an equivalence cluster. 
This can be achieved for example with some kind of clustering: co-reference chains~\cite{Ng2002}, tree traversal~\cite{Luo2004} or graph-cut algorithms~\cite{Nicolae2006}. 
The similarity between two mentions is computed using different features. 
A recent overview~\cite{Stoyanov2009} for example, mentions 63 different features, and divides them into \textit{lexical} (string-based comparison), \textit{proximity} (number of words or paragraphs between two mentions), \textit{grammatical} (based on POS, parse trees, etc) and \textit{semantic} (gender, animacy, etc).
Most of these features are based on the \textit{content} of the named entity (e.g. some additional information provided by the parser on the superficial form it takes or on the local context given by surroundings words), while the \textit{content-independent} features refer only to the close context and are always at the intra-document level.

A more global context has been used as feature for cross-document co-reference resolution.
\cite{Huang2010} uses the categories of the Open Directory Project and combines these with local context features (words in the same sentence). 
Such an approach is static by nature and no temporal information is used, which is a key feature in the domain of event detection.

A recent opposite approach~\cite{Kotov11} uses purely temporal information to solve the co-reference resolution problem: if two named entities have the same temporal profile (in other words, when their ``bursts'' of appearances are similar) and if their approximate string matching similarity is above some threshold, these named entities are claimed to refer to the same entity. Here, in addition to temporal information, we propose to introduce social and geographical contextual features. Those features are given by the clustering of news articles where each cluster of news articles is talking about one topic-coherent event, providing social, geographical and temporal co-occurrences of the entities mentioned by the articles.

Targeting the specific domain of the news-sphere, the European Media Monitor of the Joint Research Centre (JRC) of the EC has tackled the problem of multi-lingual named entity disambiguation. 
The main issue here is the correct translation of person names and the subsequent comparison between these translations. 
The similarity between two named entities is computed through the following process:  transliteration into roman script, lowercasing, name normalization, vowel removal and finally edit distance~\cite{JRCNames}. 
An additional constraint mentioned in \cite{Pouliquen06} is that, in order to merge two named entities, they have to appear in at least one identical news cluster. 
No other use of the social relationship is exploited.
The same team released recently a database~\cite{JRCNames} of different name variants for the same entity.
This has been obtained with the process described above, plus using an external co-reference resource (Wikipedia) and manual revision.
While this database is undoubtedly a valuable resource, it is tight to the news-sphere domain, has high-precision, low-recall rates and addresses only the most frequent named entities.

 %while our proposal can be used in any event-based information system.

Our work is partially inspired by the approach proposed by Bhattacharya and Getoor~\cite{Bhattacharya2005,Bhattacharya2007}.
They target the use of complex entities in databases that provide additional information (like citation databases, where co-authors are linked, and different proceedings spelling can be linked through the use of the papers published therein). 
Their idea of using relational references to the same entity as additional information for merging shares some similarity with our use of temporal and context co-occurrences of different mentions. 
Note however the difference that our entities are much simpler (basically, just a canonical string) and the strong context information that can be provided through news events.

Another specificity of our proposed approach is to iteratively process both cross-document co-reference and event detection. A comparable iterative approach~\cite{Bhattacharya2005,Bhattacharya2007} proposes a relational clustering algorithm for iteratively discovering entities by clustering references taking into account the clusters of co-occurring references. In this approach, clusters aggregate references (i.e. lexical expressions) while we are clustering documents. Their iterative approach aims at improving entity resolution task while we aim at improving both entity resolution (through cross-document co-reference task) and event detection tasks.

\smallskip

The opposed view that personal names are normally preserved between news articles of different languages and sources can be used for named entities discovery~\cite{Shinyama2004}.
A related area where a similar idea has shown to be successful is \textit{person name disambiguation}. 
The problem here is to disambiguate between two persons sharing the same name. 
The standard approach here~\cite{Pedersen04} is to cluster the close context of each occurrence of the name.

%% file: method.tex
Recall that we are jointly solving a document clustering problem -- or more precisely, an event recognition and tracking task -- and a cross-document co-reference problem. 
We adopt the extra assumption that an intra-document co-reference resolution step has been applied to the collection but, as it is the case for many state-of-the-art systems, this initial step tends to over-generate distinct named entities (several different entities are generated for the same real entity), favoring precision over recall. 
With this extra assumption, the cross-document co-reference task amounts to merge different entities into a single one, a problem that can be formulated as ``alias detection''. 

The event recognition and tracking task is solved by adapting incremental clustering techniques to deal with temporal constraints, implementing the intuitive idea that an event should emerge as a cluster of items reporting about a restricted set of actors (people, organization) interacting in a specific location during a certain time period. 
%Note, however, that we are considering here a batch case, where the goal is to organize and index the collection around events automatically extracted from the whole collection.
The intuition behind our approach is that better cross-document named entity co-reference leads to more accurate similarity measures for clustering and better event recognition leads to better cross-document named entity co-reference.
Thereby, we use the event profile of an entity as a powerful, synthetic representation of the socio-geo-temporal information related to the entity.

Our general algorithm is outlined in Alg.~\ref{alg:algorithm}, and consists of successive applications of two steps: \textit{clustering} and \textit{co-reference}. 

The clustering algorithm takes as input a set $\mathcal{D}={d_1 \dots d_n}$ of documents, threshold parameter $\tau$ and a co-reference mapping $S$.
The output $P$ is a $n \times k$ matrix of cluster assignment. Typically, a variant of the fully-incremental (or single-pass) algorithm~\cite{Rijsbergen}, described in Sect.~\ref{sec:priorart}, is used.
Although in our experiments we performed an hard assignment, our proposed method can easily be extended to a soft-clustering setting that outputs as $P[i,j]$ the probability $p(c_j|d_i)$ that document $d_i$ belongs to cluster $c_j$.
As said above, the number of clusters $k$ is not fixed but indirectly controlled by the threshold parameter $\tau$ that determines the granularity level of the final clustering.
Each document $d_i$ will be represented by three features: its bag-of-words, its bag-of-entities and its timestamp. 
The clustering algorithm then uses a similarity measures that combines these three features for each one of the documents.
In our experiments, we weighted words and entities independently with a standard `tf-idf' scheme, concatenated them into one vector and applied cosine similarity. 
Because an event is by nature limited in time, we placed an additional time constraint when comparing a document $d$  to a cluster.
Any cluster whose mean time (the average of the timestamps of the news articles composing it) exceeds 12 days from the timestamp of $d$ is filtered out, i.e. considered as inactive.
As pointed out by~\cite{Yang1998} this is done not only for computational efficiency, but also because similar articles separated in time are less likely to belong to the same event.

The co-reference mapping $S$ decides which entities will be considered as co-refering. 
Its formal definition will be given later. It is used in the clustering algorithm as an inner entity-to-entity (or feature-to-feature) similarity matrix (like in the Generalized Vector Space model). More specifically, the $S$ matrix is a matrix representing equivalence relations between features (entities) as discovered by the cross document named entity co-reference module.

Let us now turn to the cross document named entity co-reference task. 
The \textit{coreference} algorithm takes as input the set of $\mathcal{E}=\{e_1, \dots, e_m\}$ entities and a clustering $P$ of the document set $\mathcal{D}$, in addition to two parameters $\alpha$ and $\theta$.
In our experiments, we take a pairwise approach where we compute a similarity between entities and decide to merge entities $e_i$ and $e_j$ if $\text{sim}(e_i, e_j) \geq \theta$.
This similarity is a convex linear combination between a \textit{content} and a \textit{context} similarity. 

The \textit{content} similarity only considers the canonical strings of the entities.
Each entity $e$ is represented through a sparse bag-of-words vector $w(e)$ with a non-zero entry for word $v$ if the string representing entity $e$ contains $v$.
We weight this again by a `tf-idf' scheme, where the \textit{idf} part here is the inverse of the number of distinct entities in which the word appears.
As it is desirable to take into account some forms of fuzzy matching between the constitutive words (misspellings, transliterations, \textit{etc.}), we also introduce a word-to-word similarity matrix $Y$, so that the content similarity measure between two entities $e_i$ and $e_j$ will be: 
\begin{equation}
\text{sim}_\text{content}(e_i,e_j) = w(e_i)^T Y w(e_j)
\end{equation}
where $w(e_i)$ and $w(e_j)$ are $L2$-normalized.
The matrix $Y$ is derived here from the weighted-edit distance between two words.
We learnt the weights of the edit distance using an external resource, namely the \textit{JRC-Names} list of equivalent names, after alignment of their constitutive words\footnote{see \url{http://langtech.jrc.it/JRC-Names.html}}.

While the content similarity is fixed for two given entities, their \textit{context} similarity varies depending on the clustering output. 
For each entity $e$ we takes its event-profile, namely the binary vector $c(e)$ of size $k$ whose $i^{th}$ component is one if $e$ appears in at least one document belonging the cluster $i$.
Therefore,
\begin{equation}
\text{sim}_\text{context}(e_i,e_j) = c(e_i)^T c(e_j)
\end{equation}
with $c(e_i)$ and $c(e_j)$ being  $L2$-normalized.

The final similarity is then a linear combination (late fusion) of these two measures:
\begin{equation}
sim(e_i,e_j) = \alpha\, \text{sim}_\text{content}(e_i,e_j) + (1 - \alpha)\, \text{sim}_\text{context}(e_i,e_j)
\end{equation}

Recall that, in our case, the cross document named entity co-reference amounts to merge some entities together, as the intra document co-reference solver tends to be too specific. 
We therefore compute $sim(e_i,e_j)$ for all pairs of entities and merge those whose similarity is above $\theta$.
We then take the transitive closure of this relationship, obtaining the $m \times m$ symmetric matrix $S$ with $1$ at position $S[i,j]$ if $e_i$ is merged with $e_j$.
$S$ is then used as parameter for the $\textit{clustering}$ module, controlling which of the entities are being considered as being equal.
% 
% We will adopt here a single-linkage hierarchical clustering approach, whose inputs are the similarity matrix between each pair of entities and a threshold. Let us designate by $G$ the matrix of cluster assignment of this hard hierarchical clustering algorithm. So, the clustering algorithm could be materialized by the function $G = \textit{Entity\_Clustering}(X_w, X_c, Y, \alpha, \theta')$ where $X_w$ designates the $E \times W$ matrix formed by the $E$ vectors $x_w(e)$, $X_c$  the $E \times C$ matrix formed by the $E$ vectors $x_c(e)$, $Y$ is the (fixed) inner word-to-word similarity matrix, $\theta'$ is a given threshold for the hierarchical clustering algorithm and $\alpha$ is the late fusion coefficient for combining the similarities of the canonical forms of the entities and the similarities of their event-profiles. The $\alpha$ coefficient was tuned using a subset of training entities for which the ground truth was known (see the experimental section). The matrix of cluster assignment could be directly transformed to a new entity-entity equivalence matrix that will be used for the $S_e$ metric matrix in the event clustering algorithm (when computing the similarity for the ``bag-of-entities'' facet): $S_e = G G^T$; this simply means that, in the event clustering algorithm, two ``bag-of-entities'' features that correspond to two entities in the same cluster (in the entity clustering algorithm) should be considered as identical.

Putting all this together, we formalize in Algorithm~\ref{alg:algorithm} our joint event recognition - cross-document co-reference algorithm, as the iterated application of two successive grouping algorithms, each one being influenced by the outcome of the previous one.

\begin{algorithm}{}
\caption{Joint Entity and Event Clustering}
\label{alg:algorithm}
 \begin{algorithmic}[1]
  \REQUIRE{a document set $\mathcal{D}$, the entity set $\mathcal{E}$, parameters $\tau, \theta$ and $\alpha$}
%  \ENSURE{a clustering $\langle C, \gamma \rangle$ over $P$}
	\STATE $S^{(0)} = I$, the identity matrix
	\STATE $t=1$
	\WHILE{convergence criterion is not satisfied,}
		\STATE $P^{(t)} = \textit{clustering}(\mathcal{D}, S^{(t-1)}, \tau)$
		\STATE $S^{(t)} = \textit{coreference}(\mathcal{E}, P^{(t)}, \alpha, \theta)$
		\STATE $t=t+1$
	\ENDWHILE
	\RETURN $P^{(t)}$ and $S^{(t)}$

 \end{algorithmic}

\end{algorithm}

%% file: experiments.tex
For the experiments, we used the TDT5 news article corpus\footnote{\url{www.ldc.upenn.edu/Projects/TDT5}}. 
This collection includes about 280,000 documents (we considered only the articles written in English). 
Each document was linguistically pre-processed by the Xerox Incremental Parser~\cite{XIP} and this pre-processing included intra-document co-reference resolution.
We limited ourselves to the following list of named entities: persons, places, organizations and linguistic events\footnote{contrary to what we call "event" in this paper, a linguistic event is a lexical expression referring to an event such as "London Olympic Games" }.
%A trivial cross-document co-reference was then performed by merging named entities that have the same canonical string.
We represent each document by the concatenation of the tf-idf vectors of its named entities and bag-of-words (early fusion) or of its named entities alone (see below).

We evaluated the clustering using the TDT5 ground truth. 
In this dataset, 6364 articles are annotated with 126 events (called stories or topics in TDT5), which we took as gold reference for assessing the clustering performance. The clustering algorithm is the fully-incremental one, with time constraints (``old'' clusters are filtered out) and including the entity co-reference mapping as explained in Section \ref{sec:corefer}. Note that, while the algorithms are applied on the 280,000 documents, for assessing the clustering performance we extracted the subset corresponding to the 6364 labeled documents.
We report micro-average precision and recall; adopting, as usual, the mapping between identified clusters and reference events that maximized the $F_1$ measure. 
% In order to evaluate the cross-document co-reference module, we annotated you some of the named entities corresponding to these stories.
% We took 56 of these stories and analyzed the named entities that were detected inside articles of these stories. 
% There were in total 4516 different named entities.
% Those pairs that were referring to the same underlying concepts were marked as such and constitute our ground truth.
% Note that we did not consider pairs which only appeared in disjoint stories.
% After this process we ended up with 591 equivalence classes that contained more than one expression of a named entity. In the supplementary material we provide examples.

%In each series, we run the clustering with three different threshold $\theta$ for the fully-incremental algorithm ($0.1, 0.15$ and $0.2$).
%The results are in Table~\ref{tb:clustering}.
We fixed $\tau=0.1$, a value that leads to the best results for the clustering at the first iteration (which will be used as a baseline).

\begin{figure}
 \includegraphics[width=.55\textwidth]{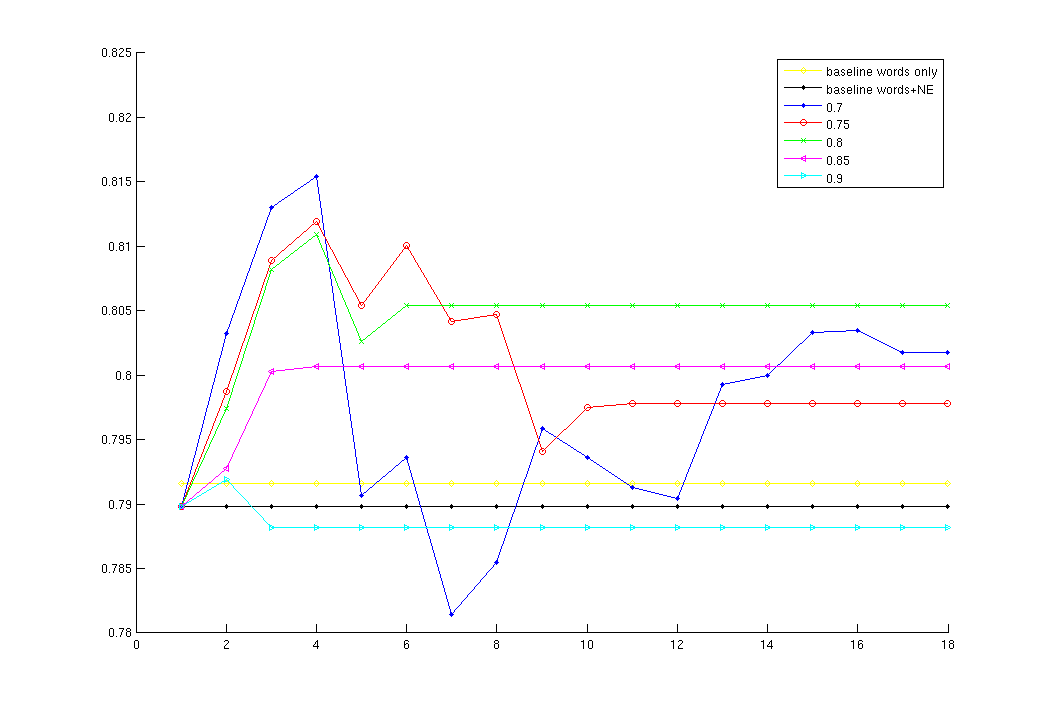}
 \caption{Evolution of the clustering ($F_1$) over successive iterations using words+NE}
 \label{fig:wordsNE}
\end{figure}

In Figure~\ref{fig:wordsNE} we plotted the evolution of the micro-average $F_1$ measure of the clustering over the different iterations. 
We ran this experiment with different values of $\theta$ and used two baselines:

\begin{enumerate}
 \item The clustering after the first iteration
 \item The clustering using words only (without considering named entities).
\end{enumerate}
 
For the co-reference module, we fixed $\alpha$ manually to $0.75$, a value which has not been optimized further.
In order to see the impact of the social similarity, we run the experiments with $\alpha=1$.
In general, the results were slightly better than those obtained with $\alpha=0.75$ at the first round, but worse than the results after convergence.
With a value of $\alpha=1$, it does not make sense to use a bootstrapping cycle like the one depicted in Figure~\ref{fig:outline}, as the co-reference decision is independent of the current clustering. 

% This was done to see better the impact of the co-reference module on the clustering: when taking into consideration words plus named entities it may well be the case that merging some of the named entities would not have a significant impact on the overall similarity.

Considering Figure~\ref{fig:wordsNE}, a first observation to note is the fact that the baseline using only the bag-of-words as features performs (slightly) better than the case where this is combined with the bag-of-entities.

However, after a couple of iterations improving the co-reference resolution, the clustering results also improve, emphasizing the importance of having a good entity resolution.
As can been seen in this figure, taking a value of $\theta$ (the threshold for the co-reference module) between $0.75$ and $0.85$ improves the clustering in a consistent way.
Outside of this range, the evolution is unstable, or even negative.
Indeed, with a lower value of $\theta$, the co-reference component precision decreases, introducing errors that affect negatively the next clustering iterations. 
With a too high value of $\theta$, fewer referents are marked as being coreferring (reducing recall) and the changes are not significant enough to affect the clustering results.

\begin{figure}
 \includegraphics[width=.55\textwidth]{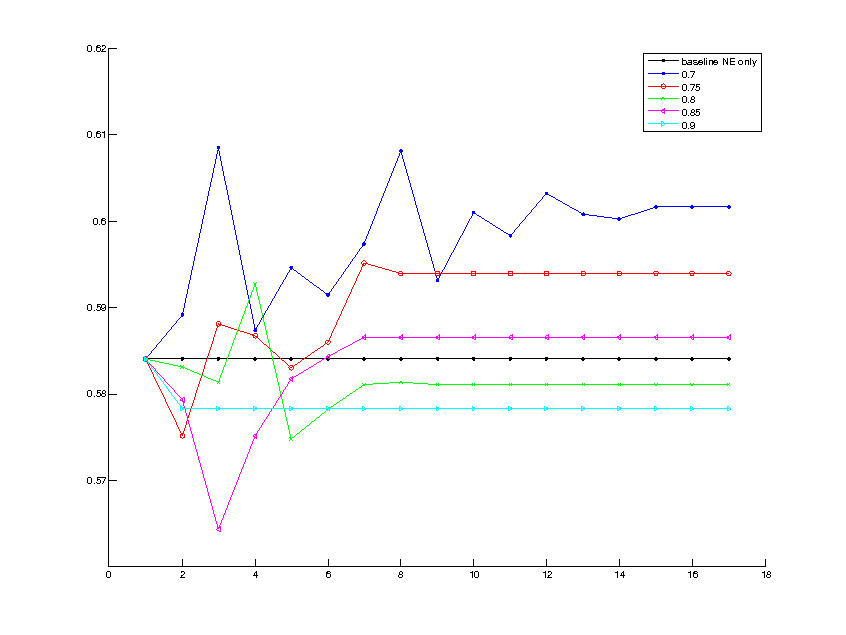}
 \caption{Evolution of the clustering ($F_1$) over successive iterations using NE only}
 \label{fig:onlyNE}
\end{figure}

In order to further investigate the connections between having a better co-reference resolution of entities and the clustering quality, we represented each document as the vector of its named entities alone, instead of combining the words and the named entities.
The corresponding results on the clustering are plotted in Figure~\ref{fig:onlyNE}.
As expected, event detection performance decreases as the entities alone do not seem to carry enough information to cluster correctly the documents.
While some of the previous conclusions also hold in this case (a general improvement), the results seem to be much more sensitive to the chosen value of $\theta$. 
This seems to indicate that the output of the co-reference module is not very stable, which can be explained by seeing the low values for the clustering ($F_1=0.56$ is a rather low value for a clustering).
The fact that a relatively bad input produces inconsistent results should not come as a surprise.

\smallskip

Until now, we only evaluated the co-reference of named entities indirectly, through its impact on the clustering.
In order to evaluate the opposite (the impact of the clustering on the co-reference resolution) directly, we created our own gold standard.
From the 126 events provided by the TDT5 ground truth, we took randomly 56 of these events and analyzed the named entities that were detected inside articles of these events. 
There were in total 4516 different named entities.
We manually annotated all the reference pairs that refer to the same real entity.
Note that, in order to reduce the amount of comparison, we considered only pairs that appeared in the same event.
Those annotated pairs that were referring to the same underlying entity constituted our cross-document named entity co-reference ground truth.
This resulted in 591 equivalence classes that contained more than one named entity. 
The fact of considering only entities that co-occur in at least one of the selected events is less than optimal: we do not considered this to be a clean gold standard, but rather a sanity check to see if the performance of both tasks are effectively correlated.

\begin{figure}
 \includegraphics[width=.55\textwidth]{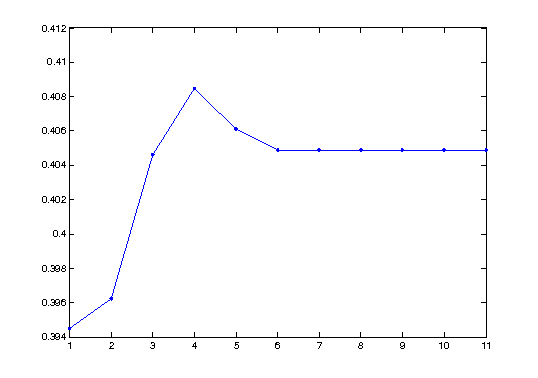}
 \caption{Evolution of the cross-document co-reference ($F_1$) over successive iterations for $\theta=0.8$.}
 \label{fig:necoref}
\end{figure}

In Figure~\ref{fig:necoref} we plotted the evolution of the micro-average $F_1$ measure of successive results of the co-reference module, with respect to our annotated ground truth, using $\theta=0.8$ during the process, the threshold that obtained the best result for the clustering.
Note that the values plotted there correspond actually to the threshold that achieved best $F_1$ (this was always around $0.55$): the reason why we used a larger threshold in the actual process was to introduce fewer false positives, in exchange of introducing fewer changes in general.
As can be appreciated, the shape of the figure mirrors the corresponding curve in Figure~\ref{fig:wordsNE}. 
This illustrate that for each iteration, the clustering is strongly linked to the co-reference resolution result, adding evidence to our initial assumption that entity resolution and event detection are effectively correlated.
At the same time, the general low performance achieved, shows that the idea can still be improved. 
In particular, considering the cross-document co-reference as an ``alias detection task'' works for the first iterations but after the fourth iteration it seems to create too many aliasing with the consequence of decreasing the result quality. % (see the following Section for further discussion on this point).

%% file: conclusions.tex
% The most striking conclusion of the results from last section seems to be the difference in behavior between the two series.
% In the case of considering named entities alone there is an improvement in both tasks, which is particularly important for the case $\tau=0.1$. 
% The clustering breaks the $0.6$ barrier for $F_1$, and the entity resolution approaches the values of considering also the bag-of-words.
% In the series of considering also the words however the results remains mostly inalterable.
% This is probably due to the fact that the consequences of merging some of the named entities are not strong enough to vary significantly the clustering.
% We did not try reducing the threshold $\theta'$ (in order to merge more entities) for this, and for doing so it would probably be necessary to add more and more complex features in order to avoid too many errors.
% Although this involves learning more parameters, the encouraging results when considering named entities alone indicate the potential value of such an approach.

We considered two apparently independent problems in a larger application (a news aggregator) and tackled them jointly.
We adapted the algorithms used for the clustering of events and for co-referencing resolution of named entities to allow as additional input the output of the other algorithm.
This feedback procedure can then be used in a bootstrap fashion to improve both tasks simultaneously.
Such a general bootstrapping approach is of general interest and may very probably be applied to a wide range of applications for data mining on the web.
%For the more general case, the bulk of the document representation is in the bag-of-words, which is not modified by our system.
% It seems however clear that careful improvement in the cross-document module (by adding more features) -- which could permit to keep similar precision by improving recall -- would bring the same benefits also in this case.

\smallskip

In the experiments performed for this paper we limited the number of features for the co-reference resolution.
The evaluation results for the co-reference module (remind that our ground truth was more of a guideline) were rather low, but still had a positive impact on the resulting clustering.
State-of-the-art co-reference systems use up to 60 different features (see Sect.~\ref{sec:priorart}): adding some of those and weighting them properly should help the co-reference resolution performance, and by extension, the clustering.
The point of this paper is not to present a state-of-the-art co-reference module, but to show the improvement that adding social features can have in such a task.
In order to reduce the number of parameters we only used one content feature, but the literature in this field confirms that choosing linguistically richer features could potentially improve the co-reference.
In particular we did not add additional features for acronyms, which are recurrent in the news-sphere.
Of course, this would need a proper (supervised) learning algorithm to learn the parameters, as for now we fixed the only parameter ($\alpha$) by hand.

Related to this, our general approach for co-referencing entities is based on pairwise similarities.
Previous studies have shown that directly considering a chain structure~\cite{Ng2002} or other richer structures~\cite{Luo2004,Nicolae2006} improves the result.
One advantage of our proposed method is that any feature-based co-reference module can be used, including those we just mentioned.
By just adding the context similarity as an additional feature to the mentioned methods, previous work could be plugged into our framework.

%Related to this, the threshold estimation for the co-reference module deserves more study: the current one ($0.7$) was fixed in order to give more preference of precision over recall and we did not try to optimize it further.

%Fixing the threshold for the clustering is not an easy task, and it is mostly done using an annotated testset~\cite{Allan2004}.
%In our proposed system we add an extra layer of complexity: the fact that the vectorial representation of the documents is ``cleaner'' in successive iteration motivate a dynamic threshold (that increases by a value $\epsilon$ at each iteration) which could prove useful.

Our cross-document co-reference module corrects errors where two expressions point to the same entity but does not consider cases of ambiguous entities. 
Moreover, we assumed that the canonical forms returned by the initial co-reference detection engine are as specific as possible.
While our experience showed us that this is a valid assumption in most cases, there are cases where the expressions are not specific enough. 
Obvious cases in the news sphere (where new entities are continually appearing) are of course people who share exactly the same name, but other cases can happen also.
Take as example an article where all references of ``New York'' are actually referring to the city of New York. 
Merging this with the expression ``New York City'' would be correct in this context, but wrong in a context where this same string refers to the US state instead of the city.
We believe that this could explain a strange phenomenon that appears in Figure~\ref{fig:wordsNE} and~\ref{fig:necoref}: a sharp jump in the first iterations, followed by a drop and a slower ascent.
This may be due to the fact that at the first iterations we do correct most of the errors due to overspecifity, but then perform more errors by merging entities that do not point to the same real entity.
An improvement to this would be a system where the merging of expressions would be done in a context-specific way, instead of globally. 
Each event may have a co-reference table that indicates which expression lead to the same entity \textit{in this context} and new documents will be compared using that table.
This does however underuses the available information by restricting only to the context where enough evidence is found.
It would be desirable to define a notion of similar contexts in order to generalize the obtained information.
Related to this is the consideration of a flexible threshold $\theta$ that changes at each iteration.